\documentclass[runningheads]{llncs}

\usepackage{graphicx}

\usepackage{amsmath,amsfonts}
\usepackage{mathtools}

\usepackage{booktabs}

\newcommand{\e}[1]{\emph{#1}}

\newcommand{\mb}[1]{\mathbb{#1}}

\newcommand{\sbeq}{\subseteq}

\newcommand{\ol}[1]{\overline{#1}}
\newcommand{\ul}[1]{\underline{#1}}

\newcommand{\R}{\mb{R}}

\begin{document}

\title{Classifying token frequencies using angular Minkowski $p$-distance}

\author{Oliver Urs Lenz\orcidID{0000-0001-9925-9482} \and
Chris Cornelis\orcidID{0000-0002-7854-6025}}
\authorrunning{O. U. Lenz \& C. Cornelis}

\institute{Research Group for Computational Web Intelligence,\\Department of Applied Mathematics, Computer Science and Statistics,\\
Ghent University
\email{\{oliver.lenz,chris.cornelis\}@ugent.be}}
\maketitle           
\begin{abstract}
Angular Minkowski $p$-distance is a dissimilarity measure that is obtained by replacing Euclidean distance in the definition of cosine dissimilarity with other Minkowski $p$-distances. Cosine dissimilarity is frequently used with datasets containing token frequencies, and angular Minkowski $p$-distance may potentially be an even better choice for certain tasks. In a case study based on the \e{20-newsgroups} dataset, we evaluate clasification performance for classical weighted nearest neighbours, as well as fuzzy rough nearest neighbours. In addition, we analyse the relationship between the hyperparameter $p$, the dimensionality $m$ of the dataset, the number of neighbours $k$, the choice of weights and the choice of classifier. We conclude that it is possible to obtain substantially higher classification performance with angular Minkowski $p$-distance with suitable values for $p$ than with classical cosine dissimilarity.

\end{abstract}

\section{Introduction}

Cosine (dis)similarity \cite{rosner56new,salton62some} is a popular measure for data that can be characterised by a collection of token frequencies, such as texts, because it only takes into account the relative frequency of each token. Cosine dissimilarity is particularly relevant for distance-based algorithms like classical (weighted) nearest neighbours (NN) and fuzzy rough nearest neighbours (FRNN). In the latter case, cosine dissimilarity has been used to detect emotions, hate speech and irony in tweets \cite{kaminska23fuzzy}.

A common way to calculate cosine dissimilarity is to normalise each record (consisting of a number of frequencies) by dividing it by its Euclidean norm, and then considering the squared Euclidean distance between normalised records. Euclidean distance can be seen as a special case of a larger family of Minkowski $p$-distances (namely the case $p = 2$). It has previously been argued that in high-dimensional spaces, classification performance can be improved by using Minkowski $p$-distance with fractional values for $p$ between 0 and 1 \cite{aggarwal01surprising}.

In light of this, we propose \e{angular Minkowski $p$-distance}: a natural generalisation of cosine dissimilarity obtained by substituting other Minkowski $p$-distances into its definition. The present paper is a case study of angular Min\-kowski $p$-distance using the well-known \e{20-newsgroups} classification dataset. In particular, we investigate the relationship between the hyperparameter $p$, the dimensionality $m$, the number of neighbours $k$, and the choice of classification algorithm and weights.

To the best of our knowledge, this topic has only been touched upon once before in the literature. Unlike the present paper, the authors of \cite{france12distance} do not evaluate classification performance directly, but rather the more abstract notion of `neighbourhood homogeneity', and they only consider a limited number of values for $p$ and $m$.
 
The remainder of this paper is organised as follows. In Section~\ref{sec_angular_minkowksi_p_distance}, we motivate and define angular Minkowski $p$-distance. In Section~\ref{sec_nn_frnn}, we recall the definitions of NN and FRNN classification. Then, in Section~\ref{sec_experimental_setup}, we describe our experiment, and in Section~\ref{sec_results} we present and analyse our results, before concluding in Section~\ref{sec_conclusion}.

\section{Angular Minkowski $p$-distance}
\label{sec_angular_minkowksi_p_distance}
In this section, we will work in a general $m$-dimensional real vector space $\mb{R}^m$, for some $m \in \mb{N}$.

The cosine similarity between any two points $x, y \in \mb{R}^m$ is defined as the cosine of the angle $\theta$ between $x$ and $y$. We obtain the cosine dissimilarity by subtracting the cosine similarity from 1. Defined thus, cosine similarity and dissimilarity take values in, respectively, $[-1, 1]$ and $[0, 2]$. However, when all records are located in $\mb{R}_{\geq 0}^m$, such as token frequencies, both measures take values in $[0, 1]$.

It is a well-known fact that cosine dissimilarity is proportional to the squared Euclidean distance between $x$ and $y$ once these points have been normalised by their Euclidean norm (note that $\cdot$ denotes the vector in-product): 

\begin{equation}
\label{eq_cosine}
\begin{aligned}
 1 - \cos \theta &= 1 - \frac{x \cdot y}{\left\vert x \right \vert \left\vert y \right \vert}\\
 &= 1 - \frac{x}{\left\vert x \right \vert} \cdot \frac{y}{\left\vert y \right \vert}\\
 &= \frac{1}{2}\left\vert\frac{x}{\left\vert x \right\vert}\right\vert + \frac{1}{2}\left\vert\frac{y}{\left\vert y \right\vert}\right\vert - \frac{x}{\left\vert x \right \vert} \cdot \frac{y}{\left\vert y \right \vert}\\
 &= \frac{1}{2} \left(\frac{x}{\left\vert x \right \vert} \cdot \frac{x}{\left\vert x \right \vert} + \frac{y}{\left\vert y \right \vert} \cdot \frac{y}{\left\vert y \right \vert} - 2\frac{x}{\left\vert x \right \vert} \cdot \frac{y}{\left\vert y \right \vert}\right)\\
 &= \frac{1}{2} \left(\frac{x}{\left\vert x \right \vert} - \frac{y}{\left\vert y \right \vert}\right)^2\\
\end{aligned}
\end{equation}

The Euclidean norm is the special case $p = 2$ of the more general Minkowski $p$-size, defined for any $x \in \mb{R}^m$ as:

\begin{equation}
\label{eq_rooted_minkowski}
\begin{aligned}
 \left\lvert x \right\rvert_p &= \left(\sum \left\lvert x_i^p \right\rvert \right)^{\frac{1}{p}},\\
\end{aligned}
\end{equation}
where $p$ is allowed to be any positive real number. Note that this is only a norm for $p \geq 1$. The Minkowski $p$-distance between any two $x, y \in \mb{R}^m$ is defined as the $p$-size of their difference $\left\lvert y - x \right\rvert_p$. This is a metric if $p \geq 1$.

Similarly, we can also view the squared Euclidean norm (distance) as the special case $p = 2$ of the \e{rootless} Minkowski $p$-size (distance), defined for any $x \in \mb{R}^m$ as:

\begin{equation}
\label{eq_rootless_minkowski}
\begin{aligned}
 \left\lvert x \right\rvert_p^p &= \sum \left\lvert x_i^p \right\rvert,\\
\end{aligned}
\end{equation}
The rootless $p$-size is not a norm for any $p$ (other than $p = 1$, for which it coincides with the ordinary $1$-norm); rootless $p$-distance is a metric for $p \leq 1$.

With these definitions in place, we can define the angular Minkowski $p$-distance between any two vectors $x, y \in \R^m$ as:

\begin{equation}
\label{eq_minkowski_p_distance}
\begin{aligned}
 \left\lvert \frac{y}{\left\lvert y \right\rvert_p} - \frac{x}{\left\lvert x \right\rvert_p} \right\rvert_p.\\
\end{aligned}
\end{equation}
as well as their rootless angular Minkowski $p$-distance:

\begin{equation}
\label{eq_rootless_minkowski_p_distance}
\begin{aligned}
 \left\lvert \frac{y}{\left\lvert y \right\rvert_p} - \frac{x}{\left\lvert x \right\rvert_p} \right\rvert_p^p.\\
\end{aligned}
\end{equation}
Thus, cosine dissimilarity corresponds to rootless angular Minkowski 2-distance, and we can consider angular Minkowski $p$-distance with different values for $p$ as alternatives to cosine dissimilarity.

\section{Classical and fuzzy rough nearest neighbour classification}
\label{sec_nn_frnn}

We will now briefly review the definition of classical weighted nearest neighbour (NN) classification \cite{fix51discriminatory,dudani73experimental,dudani76distance} and fuzzy rough nearest neighbour classification (FRNN) \cite{jensen08new,lenz23fuzzy}. Both approaches require a choice of a dissimilarity measure, weights, and a positive integer $k$ determining the number of nearest neighbours to be considered. In what follows, we will specify the class prediction that each method makes for a test instance $y$, given a training set $X$ and a decision class $C \sbeq X$.

\subsection{Nearest neighbour classification}

For NN, let $x_i$ be the $i$th nearest neighbour of $y$ in $X$. Then the class score for $C$ is given by:

\begin{equation}
\label{eq_nn}
 \left.\sum_{i \leq k \vert x_i \in C} w_i \middle/ \sum_{i \leq k} w_i\right.
\end{equation}
where $w_i$ is the weight attributed to the $i$th nearest neighbour of $y$. Two popular choices \cite{dudani73experimental,dudani76distance} for the weights are linear distance weights:

\begin{equation}
\label{eq_nn_weights_linear}
 w_i = \begin{dcases}
        \frac{d_k - d_i}{d_k - d_1} & k > 1;\\
        1 & k = 1,\\
       \end{dcases}
\end{equation}
and reciprocally linear distance weights:

\begin{equation}
\label{eq_nn_weights_reciprocal}
 w_i = \frac{1}{d_i},
\end{equation}
where $d_i$ is the distance between $y$ and $x_i$.

\subsection{Fuzzy rough nearest neighbour classification}

Properly speaking, FRNN consists of two different classifiers, the upper and the lower approximation, which can be combined to form the mean approximation. For the upper approximation, let $d_i$ be the distance between $y$ and its $i$th nearest neighbour in $C$. Then the class score for $C$ is given by:

\begin{equation}
\label{eq_upper}
 \ol{C}(y) = \sum_{i \leq k} w_i \cdot \min(0, 1 - d_i/2).
\end{equation}
For the lower approximation, let $d_i$ be the distance between $y$ and its $i$th nearest neighbour in $X\setminus C$. Then the class score for $C$ is given by:

\begin{equation}
\label{eq_lower}
 \ul{C}(y) = \sum_{i \leq k} w_i \cdot \max(d_i/2, 1).
\end{equation}
For the mean approximation, the class score for $C$ is given by:

\begin{equation}
\label{eq_mean}
 \left(\ol{C}(y) + \ul{C}(y)\right)/2.
\end{equation}
In the definition of both the upper and the lower approximation, $\left\langle w_i \right\rangle_{i \leq k}$ is a weight vector of values in $[0, 1]$ that sum to 1. As with NN, two popular weight choices are linear weights:

\begin{equation}
\label{eq_frnn_weights_linear}
 w_i = \frac{2(k+1-i)}{k(k+1)},
\end{equation}
and reciprocally linear weights:

\begin{equation}
\label{eq_frnn_weights_reciprocal}
 w_i = \frac{1}{i\cdot \sum_{i \leq k} \frac{1}{i}}.
\end{equation}

\section{Experimental setup}
\label{sec_experimental_setup}

To evaluate angular Minkowski $p$-distance, we conduct a case study on the well known text dataset \e{20-newsgroups} \cite{joachims96probabilistic}. Originally, this contained 20\,000 usenet posts from 20 different newsgroups (1000 each) from the period February-May 1993, and was collected by Ken Lang. We use the version of this dataset provided by the Python machine learning library \e{scikit-learn} \cite{pedregosa11scikitlearn}, which comprises a training set (11\,314 records) and a test set (7532 records, consisting of later posts than those in the training set), preprocessed to remove headers, footers and quotes. 

We first convert each text into a set of words, defined as any sequence of at least two alphanumeric characters separated by non-alphanumeric characters, regardless of case. Next, we count the word frequencies per text and transform this into an $m$-dimensional dataset by selecting the top-$m$ overall most frequent words, and discarding the rest.

In order to evaluate the behaviour of NN and FRNN with angular Minkowski $p$-distance, we systematically vary different values for $p$, $m$ as well as the number of nearest neighbours $k$. In the case of FRNN, we consider the upper, lower and mean approximations separately. For both NN and FRNN, we will consider linear and reciprocally linear weights, as described in Section~\ref{sec_nn_frnn}.

For $p$, we consider all multiples of $0.1$ in the range of $[0.1, 4]$, centred on the canonical values of 1 and 2. Since $k$ and $m$ encode magnitudes, we investigate them on a logarithmic scale, with values corresponding to powers of 2 in the range of, respectively, $[1, 256]$ and $[2, 4096]$.

We measure classification performance using the area under the receiver operator characteristic (AUROC) \cite{hand01simple}.

\section{Results}
\label{sec_results}

\begin{figure*}
\centering
\includegraphics[width=\linewidth]{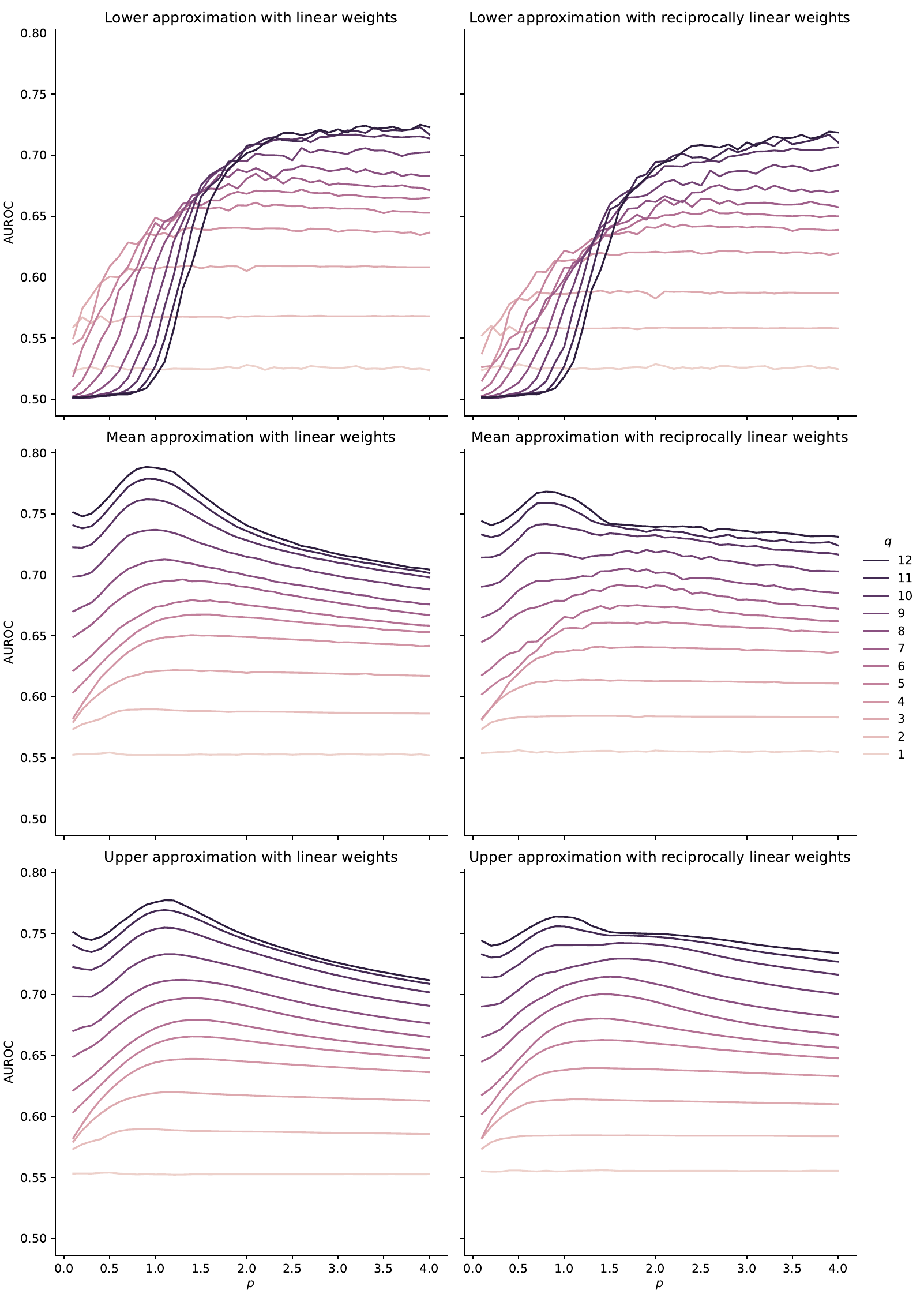}
\caption{AUROC obtained on the \e{20-newsgroups} dataset with FRNN, number of neighbours $k = 256$, dimensionality $m = 2^q$ and angular Minkowski $p$-distance.}
\label{fig_frnn_auroc}
\end{figure*}

\begin{figure*}
\centering
\includegraphics[width=\linewidth]{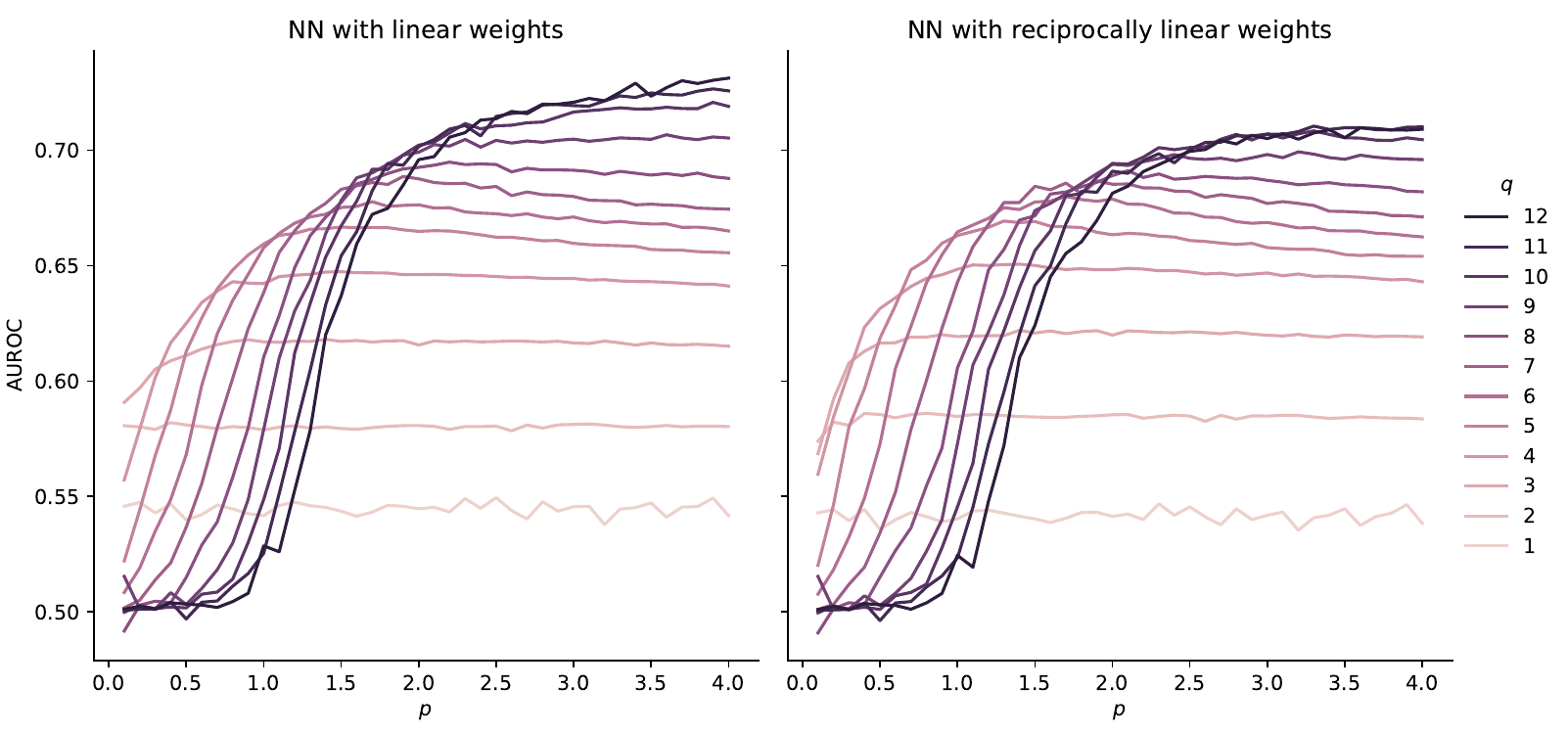}
\caption{AUROC obtained on the \e{20-newsgroups} dataset with NN, number of neighbours $k = 256$, dimensionality $m = 2^q$ and angular Minkowski $p$-distance.}
\label{fig_nn_auroc}
\end{figure*}

\begin{figure*}
\centering
\includegraphics[width=\linewidth]{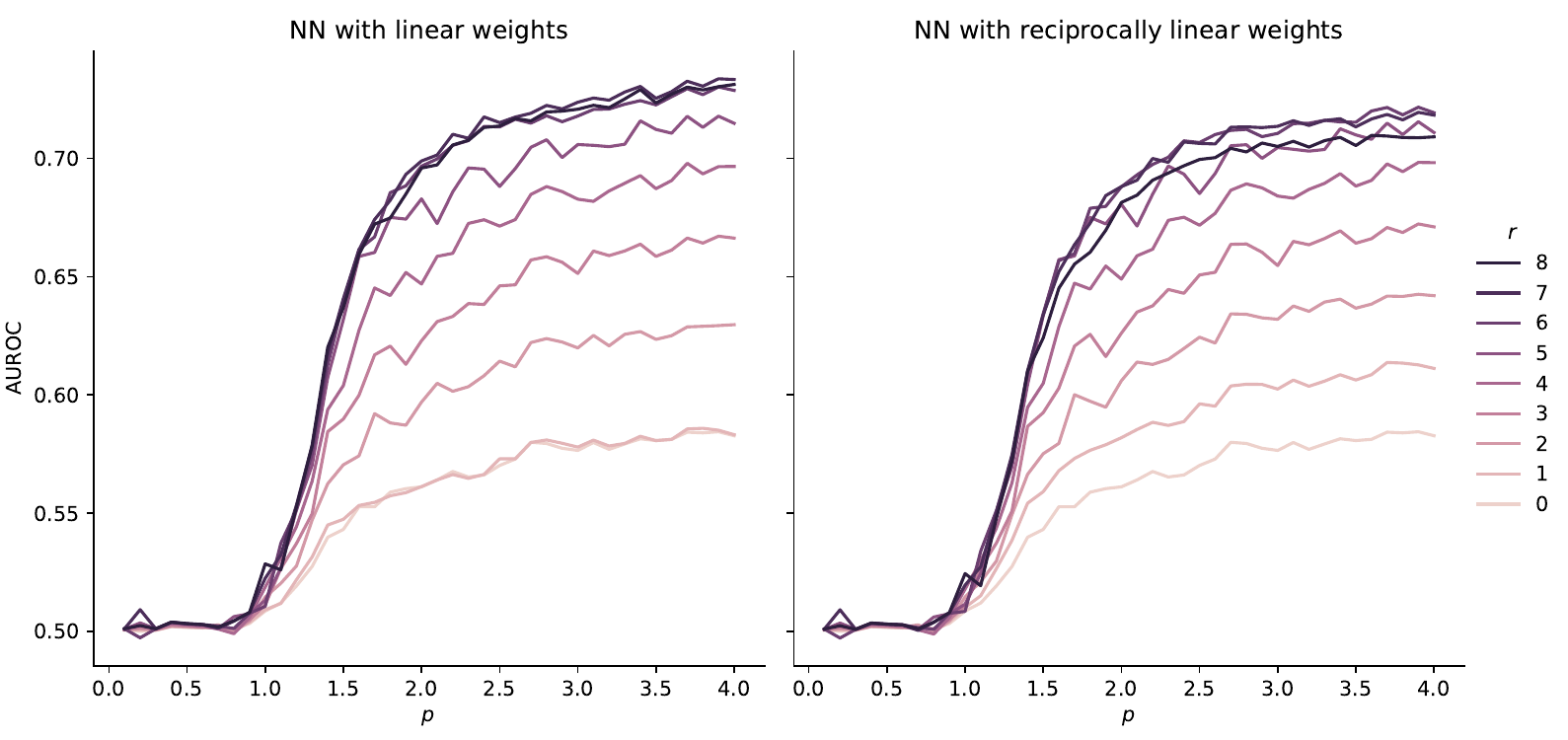}
\caption{AUROC obtained on the \e{20-newsgroups} dataset with NN, dimensionality $m = 4096$, number of neighbours $k = 2^r$ and angular Minkowski $p$-distance.}
\label{fig_nn_auroc_by_k}
\end{figure*}

\begin{figure*}
\centering
\includegraphics[width=\linewidth]{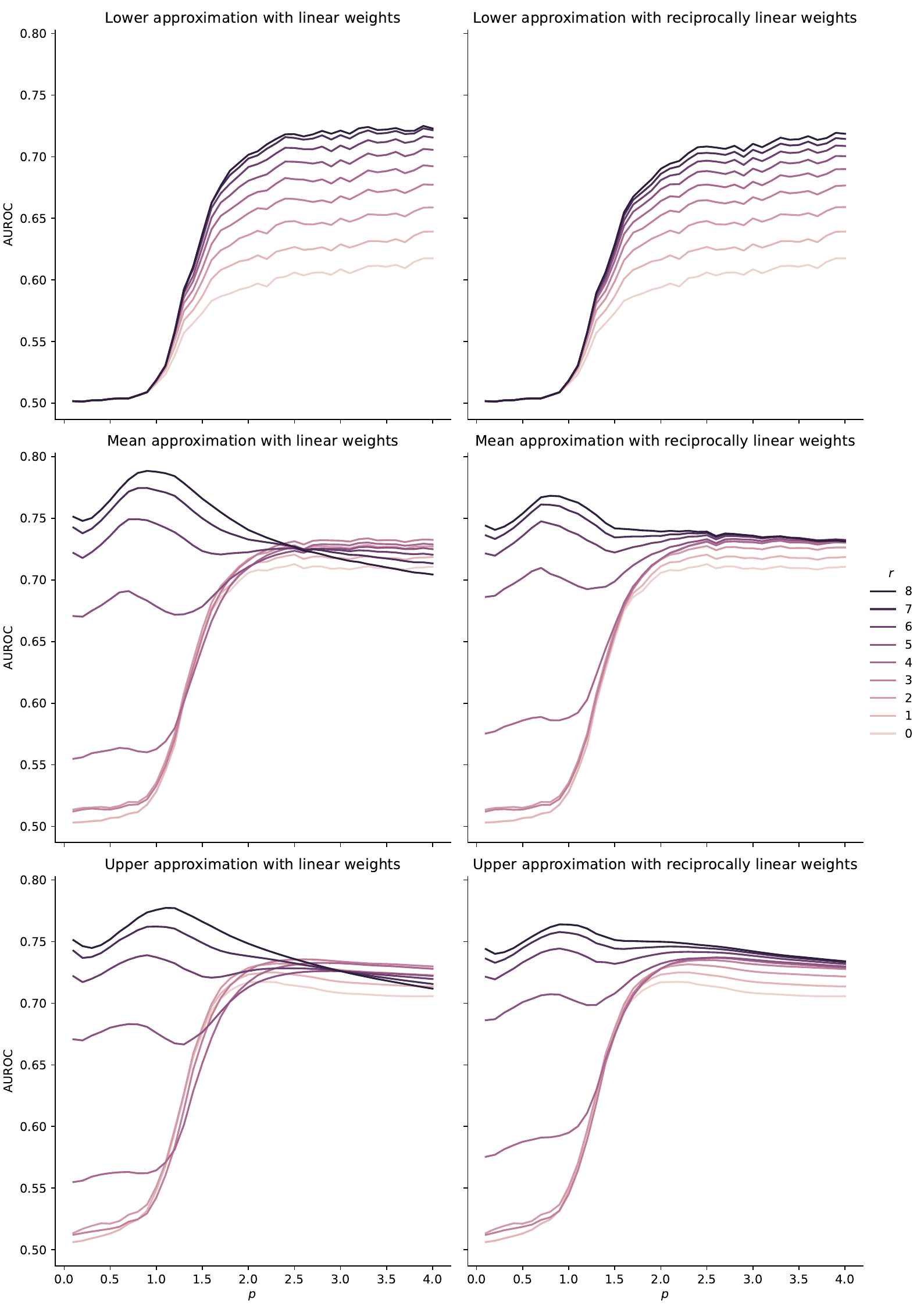}
\caption{AUROC obtained on the \e{20-newsgroups} dataset with FRNN, dimensionality $m = 4096$, number of neighbours $k = 2^r$ and angular Minkowski $p$-distance.}
\label{fig_frnn_auroc_by_k}
\end{figure*}

\begin{figure*}
\centering
\includegraphics[width=\linewidth]{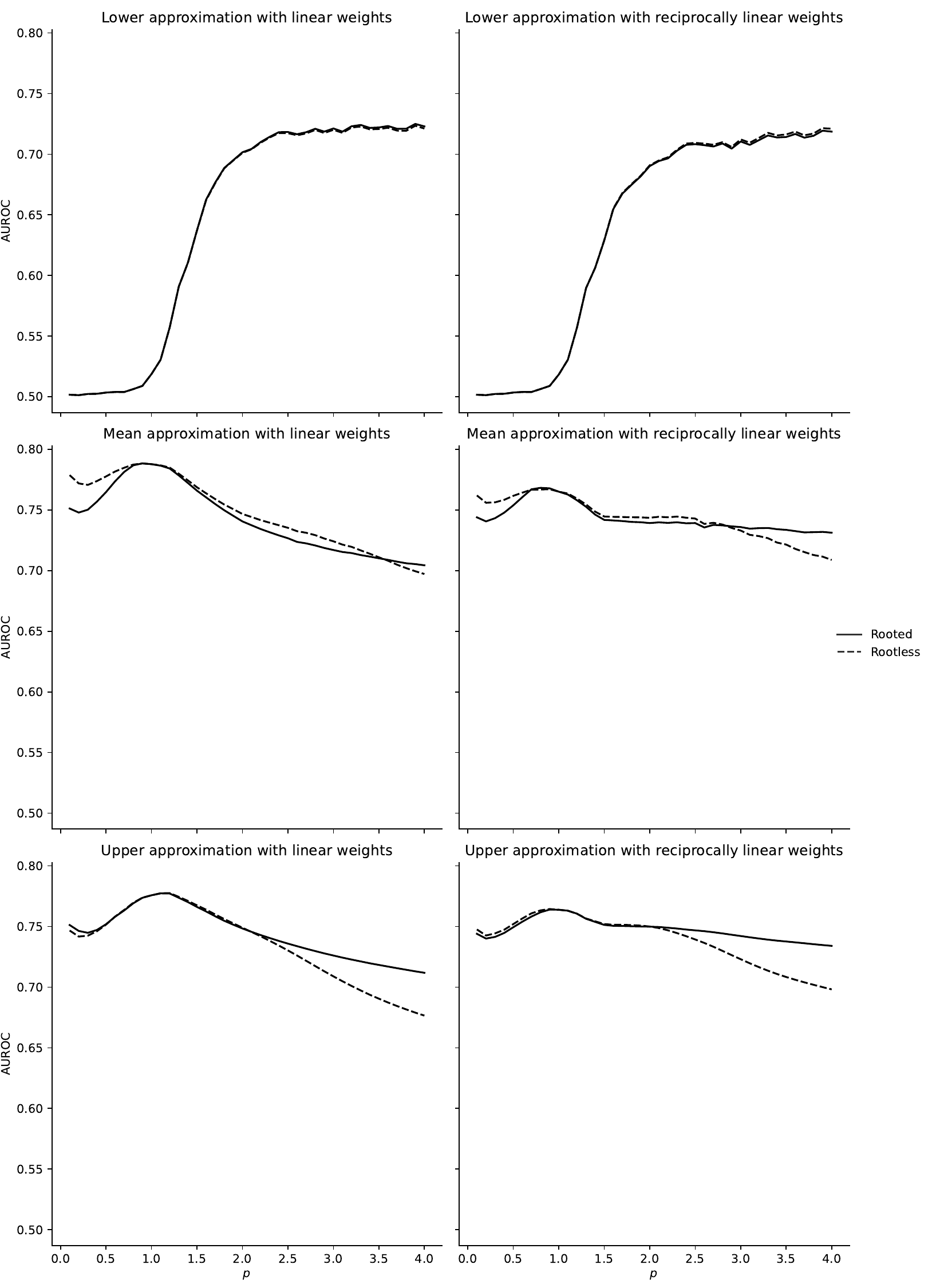}
\caption{AUROC obtained on the \e{20-newsgroups} dataset with FRNN, number of neighbours $k = 256$, dimensionality $m = 4096$ and rooted and rootless angular Minkowski $p$-distance.}
\label{fig_frnn_auroc_by_rootless}
\end{figure*}

\begin{figure*}
\centering
\includegraphics[width=\linewidth]{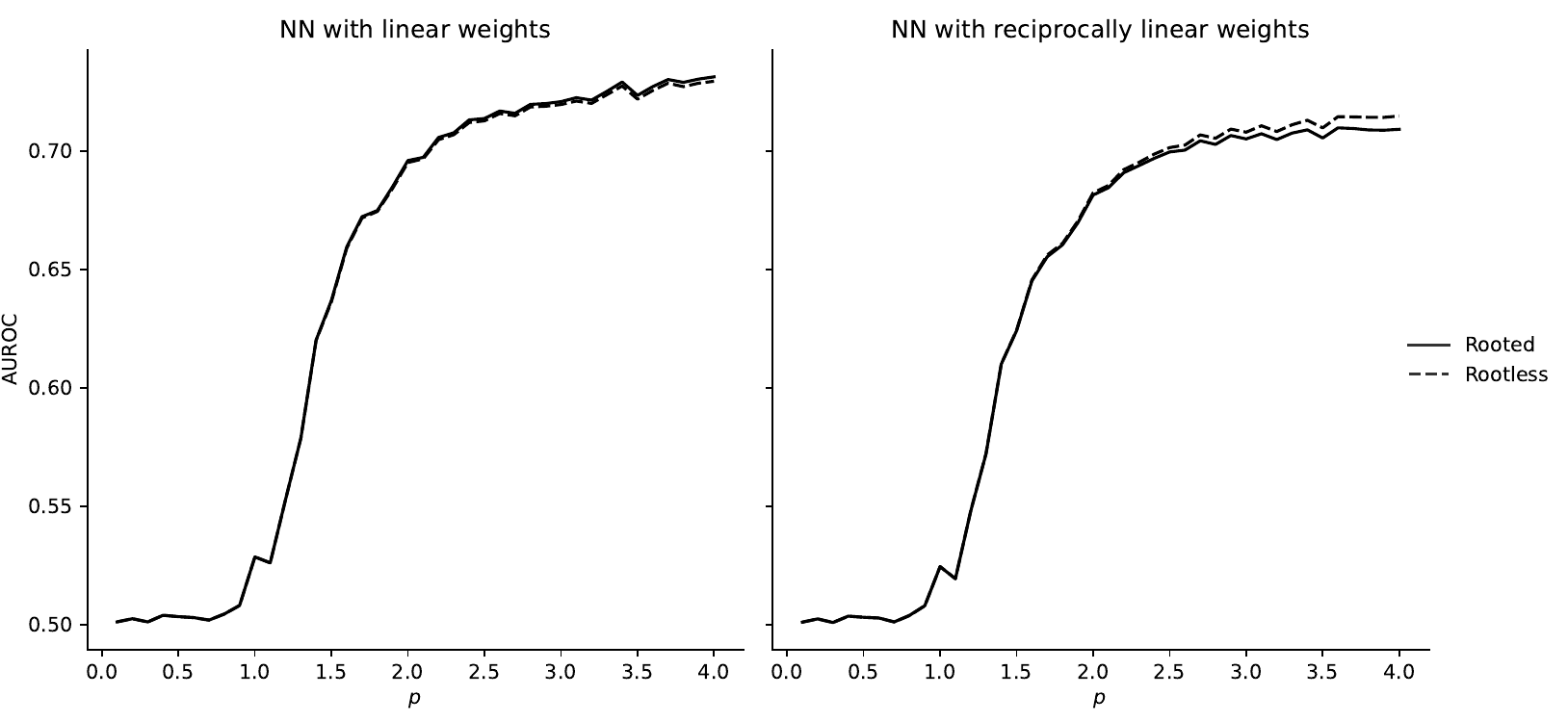}
\caption{AUROC obtained on the \e{20-newsgroups} dataset with NN, number of neighbours $k = 256$, dimensionality $m = 4096$ and rooted and rootless angular Minkowski $p$-distance.}
\label{fig_nn_auroc_by_rootless}
\end{figure*}

Figures~\ref{fig_frnn_auroc} and \ref{fig_nn_auroc} display AUROC as a function of dimensionality (the number of most frequent tokens taken into consideration) and as a function of $p$, for $k = 256$. There are a few things to be noted from these response curves:

\begin{itemize}
 \item The choice of weights doesn't appear to play a role in the overall behaviour of these response curves.
 \item The response curves are substantially smoother for the upper approximation than for the lower approximation and for NN. The mean approximation appears to inherit some of this smoothness from the upper approximation. This qualitative difference is somewhat surprising, but it can perhaps be explained by the fact that for the upper approximation, neighbours are drawn from a uniform concept (each decision class), whereas for the lower approximation and NN, neighbours are drawn from across decision classes.
 \item The upper approximation is a better classifier (in terms of AUROC) than the lower approximation and NN for the \e{20-newsgroups} dataset. Given the relatively poor performance of the lower approximation, it is surprising that the mean approximation produces even better results than the upper approximation.
 \item AUROC increases with dimensionality, but the difference between 2048 and 4096 dimensions is quite small. It appears that up until that point, the additional information encoded in each additional dimension outweighs the noise. Note, however, that even before that point, we get diminishing returns. For each subsequent curve we need to double the dimensionality, and we obtain a performance increase that is smaller than the previous one.
 \item For NN and the lower approximation, the choice for $p$ becomes more important as dimensionality increases. Not only is a good choice for $p$ necessary to make use of the potential performance increase from adding more dimensions, choosing $p$ poorly can actually cause performance to decrease with dimensionality.
 \item There is a marked difference with respect to the optimal values for $p$ between the different classifiers. For NN and the lower approximation, higher values appear to be better within the range $[0.1, 4]$ that we have investigated, albeit with diminishing returns. For the upper and mean approximations, the optimum is located near $p = 1$ for high dimensionalities. 
\end{itemize}

As mentioned above, Figures~\ref{fig_frnn_auroc} and \ref{fig_nn_auroc} reflect a choice of the number of neighbours $k = 256$. The effect of $k$ on performance is illustrated in Figures~\ref{fig_nn_auroc_by_k} and \ref{fig_frnn_auroc_by_k}, for $m = 4096$.

\begin{itemize}
 \item For NN and the lower approximation, the overall behaviour of the response curve does not change with $k$. Higher values for $k$ lead to higher AUROC, and within the range of investigated values, the relationship appears to be similar to the relationship between AUROC and $m$: each doubling of $k$ leads to an increase in AUROC that is slightly smaller than the previous increase. From $k = 128$ to $k = 256$, the increase is already quite small.
 \item In contrast, for the upper and mean approximations, AUROC starts out quite high for high values of $p$, and increases only little thereafter. Howewer, from $k = 8$ upwards, AUROC starts to strongly increase for lower values of $p$, eventually surpassing the AUROC obtained with higher values of $p$ from $k = 64$ upwards. This means that the good performance of the mean and upper approximations around $p = 1$ is only realised for high values of $k$.  
\end{itemize}

Finally, we may also ask whether it makes a difference whether we use rooted (`ordinary') or rootless angular Minkowski $p$-distance. The results discussed above were obtained using rooted angular Minkowski $p$-distance. It turns out that using rootless angular Minkowski $p$-distance, which generalises cosine dissimilarity more closely, does not make much difference (Figures~\ref{fig_frnn_auroc_by_rootless} and \ref{fig_nn_auroc_by_rootless}). In particular, there is (by definition) no difference for $p = 1$, which maximises classification performance for the upper and mean approximations.

\begin{table}
\centering
\caption{Highest AUROC and corresponding value for $p$ obtained on the \e{20-newsgroups} dataset, with linear weights, number of neighbours $k = 256$, dimensionality $m = 4096$ and rooted angular Minkowski $p$-distance.}
\label{tab_highest_auroc}
\begin{tabular}{p{.75\linewidth}p{.1\linewidth}p{.1\linewidth}}
\toprule
                Classifier & $p$ & AUROC \\
\midrule
                        NN & 4.0 & 0.731 \\
FRNN (lower approximation) & 3.9 & 0.725 \\
 FRNN (mean approximation) & 0.9 & 0.788 \\
FRNN (upper approximation) & 1.1 & 0.777 \\
\bottomrule
\end{tabular}
\end{table}

In summary (Table~\ref{tab_highest_auroc}), we obtain the best classification performance on the \e{20-news\-groups} dataset with the upper and mean approximation and angular Minkowski $p$-distance with values of $p$ around 1, but only when $k$ is high enough ($\geq 64$).

\section{Conclusion}
\label{sec_conclusion}
We have presented angular Minkowski $p$-distance, a generalisation of the popular cosine (dis)similarity measure. In an exploratory case study of the large \e{20-newsgroups} text dataset, we showed that the choice of $p$ can have a large effect on classification performance, and in particular that the right choice of $p$ can increase classification performance over cosine dissimilarity (which corresponds to $p=2$).

We have also examined the interaction between $p$ and the dimensionality $m$ of a dataset, the choice of classification algorithm (NN or FRNN), the choice of weights (linear or reciprocally linear), and the choice of the number of neighbours $k$. We found that while the choice of weights was not important, the best value for $p$ can depend on $m$, $k$ and the classification algorithm. Under optimal circumstances (high $k$ and high $m$), the best-performing values for $p$ are in the neighbourhood of $1$ (FRNN with upper or mean approximation) and around $4$ (NN and FRNN with lower approximation).

A major advantage of angular Minkowski $p$-distance is that it is defined in terms of ordinary Minkowski $p$-distance, which is widely available. Thus, angular Minkowski $p$-distance does not require any dedicated implementation and can easily be used in experiments by other researchers.

The most important open question to be investigated in future experiments is to which extent these results generalise to other text datasets, as well as to other datasets containing token frequencies. Depending on the outcome of these experiments, it may be possible to formulate more general conclusions about the best choice for $p$, or we may be forced to conclude that this is a hyperparameter that must be optimised for each individual dataset.

\subsubsection{Acknowledgements} The research reported in this paper was conducted with the financial support of the Odysseus programme of the Research Foundation -- Flanders (FWO).

\bibliographystyle{splncs04}
\bibliography{20220728_minkowski_cosine}

\end{document}